\documentclass[authoryear,review,12pt]{elsarticle}

\usepackage[margin=1in]{geometry}
\usepackage{setspace}
\usepackage{bm}
\usepackage{bbm}
\usepackage{url}
\usepackage{wrapfig}
\usepackage{graphicx}
\usepackage{subcaption}

\usepackage{amsmath}
\usepackage{amsfonts}
\usepackage{algorithmic,algorithm}

\usepackage{gensymb}
\usepackage{soul}
\usepackage{color}
\usepackage{longtable}
\usepackage{booktabs}
\usepackage{hyperref}

\DeclareMathOperator*{\simil}{sim}

\usepackage{tikz}
\renewcommand{\cite}{\citep}

\newcommand{\reasat}[1]{%
\ifproofread
\textcolor{red}{#1}%
\else
#1%
\fi
}
\newif\ifproofread

\proofreadfalse
\usepackage{amssymb}
\journal{MLWA}

\begin{document}

\begin{frontmatter}



\title{Data Efficient Contrastive Learning in Histopathology using Active Sampling}


\author[inst1]{Tahsin Reasat}
\ead{tahsin.reasat@vanderbilt.edu}
% \cortext[cor1]{Corresponding author}

\affiliation[inst1]{organization={Department of Electrical and Computer Engineering},%Department and Organization,
            addressline={Vanderbilt University}, 
            city={Nashville},
            postcode={37235}, 
            state={Tennessee},
            country={USA}}
\author[inst2]{Asif Sushmit}
\ead{sushmit@ieee.org}

\author[inst1,inst3]{David S.~Smith}
\ead{david.smith@vumc.org}

\affiliation[inst2]{organization={Bengali.AI},%Department and Organization
            addressline={West Dhanmondi Shankar}, 
            city={Dhaka},
            postcode={1215}, 
            % state={Tennessee},
            country={Bangladesh}}

            \affiliation[inst3]{organization={Institute of Imaging Science},%Department and Organization
        addressline={Vanderbilt University Medical Center}, 
            city={Nashville},
            postcode={37232}, 
            state={Tennessee},
            country={USA}}

\begin{abstract}
Deep learning (DL) based diagnostics systems can provide accurate and robust quantitative analysis in digital pathology. These algorithms require large amounts of annotated training data which is impractical in pathology due to the high resolution of histopathological images. Hence, self-supervised methods have been proposed to learn features using ad-hoc pretext tasks. The self-supervised training process uses a large unlabeled dataset which makes the learning process time consuming. In this work, we propose a new method for actively sampling informative members from the training set using a small proxy network, decreasing sample requirement by 93\% and training time by 62\% while maintaining the same performance of the traditional self-supervised learning method. \reasat{The code is available on \href{https://github.com/Reasat/data_efficient_cl}{github}.}
\end{abstract}

\iffalse
\begin{graphicalabstract}
\includegraphics{grabs}
\end{graphicalabstract}

\begin{highlights}
\item Research highlight 1
\item Research highlight 2
\end{highlights}
\fi

\end{frontmatter}


\section{Introduction}
% about self supervised learning
\reasat{Self-supervised learning (SSL) is a subset of unsupervised learning where the model is trained using automatically generated labels derived from the data itself rather than relying on manually labeled data \cite{chen2020simple}. The key idea is to create a learning signal by constructing tasks that the model can solve without human supervision thus reducing the need to have a large manually annotated data.} The recent renaissance of SSL began with artificially designed pretext
tasks, such as relative patch prediction \cite{doersch2015unsupervised}, solving jigsaw puzzles \cite{noroozi2016unsupervised}, colorization \cite{zhang2016colorful} and rotation prediction \cite{gidaris2018unsupervised}. 
Recently, contrastive learning (CL) has emerged as a simple yet promising tool in SSL \cite{chen2020simple}. 
It has been successfully applied to medical predictive tasks like cancer detection and segmentation in histopathology images \cite{ciga2022self}. 
% Good results can be obtained using big networks and longer training time \cite{kolesnikov2019revisiting}. 
% Augmentations somewhat ad-hoc heuristics, which limits the generality of learned representations.
In practice, the CL model is trained using a large unlabeled data pool, and the extracted features from this large model are used to train a smaller linear model, and its prediction is evaluated on a test set to measure its performance \cite{chen2020simple}. 

The training phase for CL is slow and learned latent space is unconstrained, which could be suboptimal for histopathological datasets. Histopathological datasets often have a lot of confounding image sections consisting of background and debris alongside normal tissue area. Additionally, the tissue of interest is normally the cancerous tumor area which is often a minority class.  As a result the model either takes too many iterations to learn the minority class examples or, even worse, ignores the minority class and uses all its capacity to learn the non-tumor cells. 
\reasat{In traditional feature extraction based machine learning, synthetic minority over-sampling technique (SMOTE) \cite{chawla2002smote}, is a popular choice to mitigate class imbalance. But this method is performed in scenarios where discriminative features can be pre-extracted from the data. But image data is high dimensional and traditional feature extractors do not capture discriminative features as well as DL models \cite{krizhevsky2012imagenet}. DL models learn features in a data-driven approach as we do not have access to well-trained informative features in the initial stage of training, it is difficult to adapt techniques like smote to tackle imabalance. However, it has been shown that in a supervised setting, DL can be combined with active learning \cite{coleman2019selection} by training small proxy models. In this work we, explore the idea of using active learning to mitigate imbalance in CL scenario.}
%segmentation literature 
% Cross-patch Dense Contrastive Learning for Semi-supervised Segmentation
% of Cellular Nuclei in Histopathologic Images
% Unsupervised Representation Learning for Tissue Segmentation in Histopathological Images: From Global to Local Contrast
% Looking Beyond Single Images for Contrastive Semantic Segmentation Learning
% regional contrastive learning, active sampling, for semantic segmentation, semisupervised, https://arxiv.org/pdf/2104.04465.pdf
% Semi-supervised Contrastive Learning for Label-efficient Medical Image Segmentation
%vLocal contrastive loss with pseudo-label based self-training for semi-supervised medical image segmentation
%Uncertainty-guided mutual consistency learning for semi-supervised medical image segmentation

\begin{figure}
    \centering
    \includegraphics[width=0.47\textwidth]{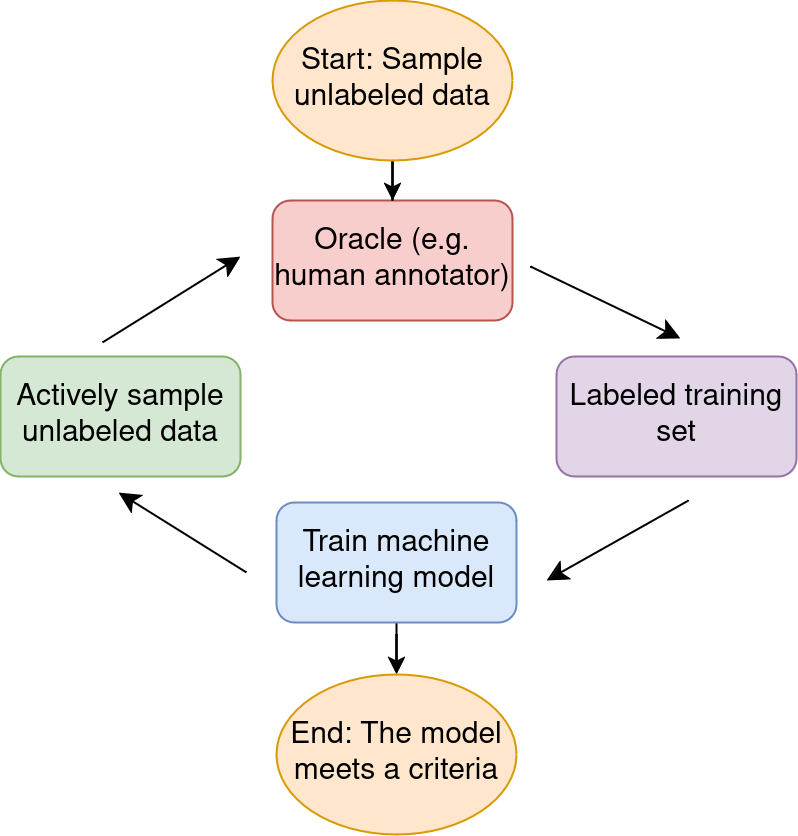}
    \caption{Active learning loop \cite{settles2009active}. An oracle annotates the most informative samples which is used to refine model predictions.}
    \label{fig:al_loop}
\end{figure}

Active learning  \cite{cohn1994improving,krogh1995neural} allows humans to interactively improve algorithms. \reasat{In Fig.~\ref{fig:al_loop}, we see the hypothetical flow diagram of a typical active learning based machine learning algortihm}. A chosen algorithm is trained iteratively using an algorithmically sampled data subset (active sampling) which is labeled by an oracle (e.g., human annotator). In each iteration,  new informative subset of samples is chosen to further train and improve model predictions. Multiple datasets have been produced by user-refined inputs such as automatic teller machine, self-driving cars, automatic content tagging in online platforms \cite{lecun2015deep}. AL has proven useful in annotating data when specialist input is required, e.g. medical images \cite{budd2019survey}. 

To decrease data redundancy, we propose to train the network iteratively using batches of unlabeled data.
The first batch is randomly selected and after training the model on each batch, we select the next batch of data from the unlabeled pool using a proxy model and a sample selection strategy and append it to the current batch. This enables the model to learn features using the most informative samples, increasing sample efficiency and decreasing time complexity. The contributions of our work are to 
\begin{itemize}
    \item increase CL efficiency in terms of time and sample complexity,
    \item provide a method to constrain the model into learning relevant representations by using active sampling, and
    \item improve CL efficiency in the histopathology domain.
\end{itemize}

\section{\reasat{Related Works}}

\reasat{Histopathology is the gold standard in medical diagnostics, particularly in the classification of tumors \cite{gurcan2009histopathological}. Histopathology images are analyzed to classify different types of tumors, each requiring precise identification to guide effective treatment plans. Researchers have developed DL based solutions to classify between in benign and malignant tumor in various organs such as breast \cite{bayramoglu2016deep_breast_cancer}, lung \cite{coudray2018classification_lung_cancer}, renal \cite{tabibu2019pan_renal_cancer}, liver \cite{sun2019deep_liver_cancer}, colon, gastric \cite{iizuka2020deep_gastric_colon_cancer} etc. DL requires large and costly manually annotated datasets to train models. Data labeling in histopathology is an intricate and time-consuming process, requiring expert pathologists to meticulously annotate images. Researchers have proposed data augmentation \cite{xue2021selective_augmentation, cicalese2020stypath_augmentation, faryna2021tailoring_augmentation},  transfer learning \cite{talo2019automated_transfer_learning}, synthetic data generation \cite{xue2019synthetic_synthesis}, SSL \cite{ciga2022self}, etc. However, data augmentation is constrained by the diversity of the source dataset. Transfer learning relies on the similarity of target and source domain while relying on large labeled dataset of a source domain. Synthetic data generation often lacks realism and the generated images resemble closely to the dataset distribution. 
Among the various techniques, SSL has emerged as a promising solution since it is free from these limitations and scales well with the addition of unlabeled data \cite{ciga2022self}. However, under data imbalance, the CL framework can learn irrelevant features and deter from its task of learning tumor features. Hence, we propose to iteratively select informative sample subset with the use of active learning and there are a very few works in this direction. Researchers in \cite{joshi2023data} have suggested the use of latent class centers which is approximated utilizing an external pretrained model and selects a subset of sample which lies near the cluster centers. 
%We do not use such a pretrained model. 
% Rather we use our CL model features for informative sample selection.
Our work differs from them in two aspects. The authors proposed the usage of a (CLIP model \cite{radford2021learning} or a Resnet \cite{he2016deep}) model to extract features and guide the data sampling approach. These models were trained on labeled images \cite{deng2009imagenet} or image - text pair scraped from the internet (CLIP). The data corpus has classes that do not cover medical entities and therefore not suitable for complicated medical data related tasks. We do not rely on any pretrained models and use a much smaller and simpler proxy model which we train on small relevant subset of labeled data. Moreover, their data selection strategy is based on the assumption that images that have similar features will be more useful for feature learning. This assumption is not valid for medical images in the presence of data imbalance. As demonstrated by our experiments the learning performance of the default contrastive learning scheme is inefficient due to the presence of redundant images that do not contribute to learning tumor features.}
% it is trained on natural images and not fit for fine-grained classification such as histopathological images \cite{radford2021learning}.
Researchers in \cite{coleman2019selection} used a small proxy network to select samples in an AL setup. But they consider AL in the supervised domain (they require an oracle for every iteration of the AL loop). 
Our work can be considered as an extension of their framework to the self-supervised domain.
Furthermore, this is the first work that considers data efficiency in CL for the domain of histopathology images. 
% add this ti lit review Self-supervised driven consistency training for annotation efficient histopathology image analysis
% Essential Subsets for Supervised Learning. There has
% been a recent body of efforts on finding the most important
% subsets for supervised learning. Empirical methods commonly rank examples from easiest to hardest—based on
% confidence, loss or gradient—and curate subsets preserving
% the hardest examples. Coleman et al. (2020) used a smaller
% trained proxy model to find the most uncertain examples to
% train a larger model. Toneva et al. (2019) selects examples
% with highest forgetting score, i.e. the number of times they
% transition from being classified correctly to incorrectly during training. Swayamdipta et al. (2020) selects examples
% with the highest variance of predictions during training. Paul
% et al. (2021) selected examples with the lowest expected gradient norm over multiple initializations. More theoretically
% motivated approaches iteratively select subsets by importance sampling based on gradient norm (Katharopoulos &
% Fleuret, 2018) or select weighted subset of examples which
% closely capture the full gradient (Mirzasoleiman et al., 2020;
% Pooladzandi et al., 2022; Killamsetty et al., 2021).
% In the active learning setup, the steps involve (Fig. \ref{fig:al_loop})
% \begin{itemize}
%     \item An unlabeled pool of data
%     \item A query selection algorithm
%     \item An oracle
%     \item A labeled training set
%     \item A machine learning model
% \end{itemize}
% Amongst these elements the query selection method has a key role in the performance of the active learning model.

\section{Methods}

\begin{figure*}[ht]
    \centering
    \begin{tabular}{cc}
        \begin{minipage}{0.35\textwidth}
            \centering
            \begin{subfigure}[b]{\linewidth}
                \centering
                \includegraphics[width=\textwidth]{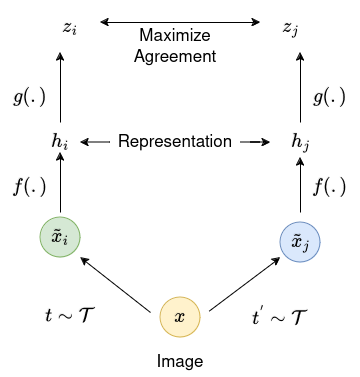}
                \caption{Contrastive learning framework from SimCLR.}
                \label{fig_simclr}
            \end{subfigure}
            \vskip\baselineskip
            \begin{subfigure}[b]{\linewidth}
                \centering
                \includegraphics[width=\textwidth]{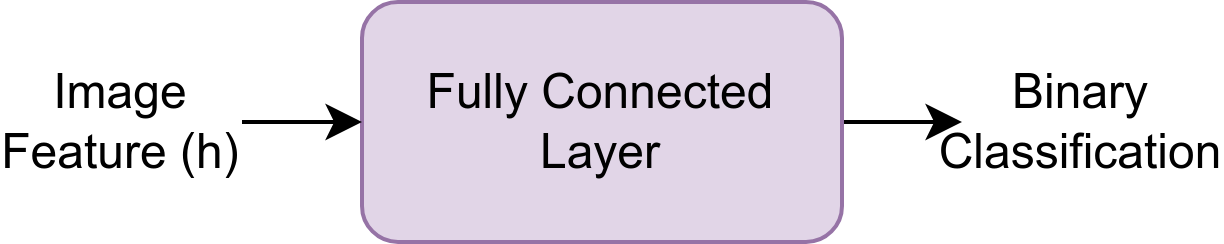}
                \caption{Proxy model.}
                \label{fig_proxy_model}
            \end{subfigure}
        \end{minipage}
        &
        \begin{minipage}{0.5\textwidth}
            \centering
            \begin{subfigure}[b]{\linewidth}
                \centering
                \includegraphics[width=\textwidth]{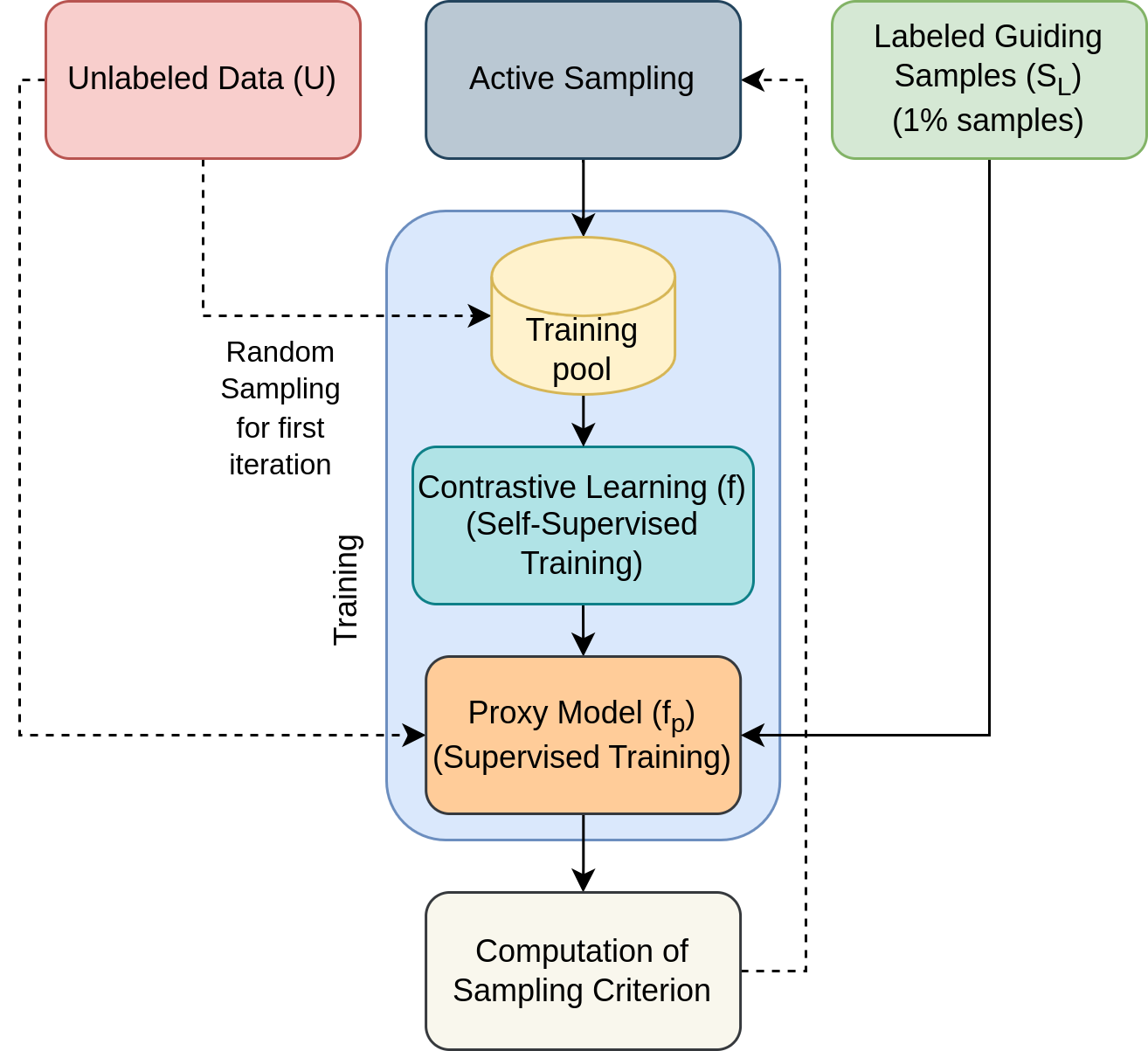}
                \caption{Proposed Framework.}
            \label{fig:proposed_framework}
            \end{subfigure}
            % \includegraphics[width=\textwidth]{fig/act_cont_framework.png}
            % \caption{Figure 3}
        \end{minipage}
    \end{tabular}
    \caption{\ref{fig_simclr}) The SimCLR framework. A neural network model $f(\cdot)$ minimizes the distance (maximizes agreement) between feature representation of two augmented views $\tilde{x}_i$ and $\tilde{x}_j$ of the same image. \ref{fig:proposed_framework}) The proposed framework speeds up the contrastive sampling process by actively selecting informative samples with the help of a simple proxy model. \ref{fig_proxy_model}) Structure of the simple proxy model which is a fully connected network with a depth of one.}
    \label{fig_models}
\end{figure*}

% \begin{figure}[ht]
%     \centering
%     \begin{subfigure}[t]{0.45\textwidth}
%         \centering
%         \includegraphics[width=\textwidth]{fig/cl_module.png}
%         \caption{Figure 1}
%     \end{subfigure}
%     \quad
%     \begin{subfigure}[t]{0.45\textwidth}
%         \centering
%         \includegraphics[height=0.9\textheight]{fig/proxy_model.png}
%         \caption{Figure 3}
%     \end{subfigure}
%     \vskip\baselineskip
%     \begin{subfigure}[t]{0.45\textwidth}
%         \centering
%         \includegraphics[width=\textwidth]{fig/proposed_framework.png}
%         \caption{Figure 2}
%     \end{subfigure}
%     \caption{Fig. with three figures: two in the first column, one in the second column}
% \end{figure}

% \begin{figure}
% \centering
%   \includegraphics[width=0.47\textwidth]{fig/SimCLR.png}
%   \caption{Fig.~2 from \cite{chen2020simple} showing the SimCLR framework. The model minimizes the distance (maximizes agreement) between feature representation of two augmented views $\tilde{x}_i$ and $\tilde{x}_j$ of the same image.}
%   \label{fig_simclr}
% \end{figure}
% \begin{figure}
%     \centering
%     \includegraphics[width=0.50\textwidth]{fig/proposed_framework.png}
%     \caption{The proposed framework speeds up the contrastive sampling process by actively selecting informative samples with the help of a simple proxy model.}
%     \label{fig:proposed_framework}
% \end{figure}
In this section, we describe the components of our proposed framework and formalize the methodology. The overview of the proposed framework is depicted in Fig.~\ref{fig_models}.

\subsection{Contrastive Learning}

SimCLR \cite{chen2020simple} proposed a simple framework for contrastive learning of visual representations by maximizing agreement between differently augmented views of the same sample via a contrastive loss in the latent space.
%Details from https://lilianweng.github.io/posts/2021-05-31-contrastive/#vision-image-embedding
\reasat{Let $U$ be a set of unlabeled image pool and $x$ are the image samples $x\in U$.} In CL, a minibatch of $N$ samples are randomly sampled and each sample is applied with random  data augmentation operations, resulting in $N$ pairs of augmented samples or $2N$ augmented samples in total:
\begin{align}
    \tilde{x}_i = t(x),\, \tilde{x}_j = t'(x)
\end{align}
where two separate data augmentation operators, $t$ and $t'$, are sampled from the same family of augmentations $\mathcal{T}$.
Given one positive pair, other  $2(N-1)$ data points are treated as negative samples. The representation is produced by a base encoder $f(\cdot)$:
\begin{align}
    h_i = f(\tilde{x}_i), \, h_j = f(\tilde{x}_j)
\end{align}
The encoded representation is passed through a projection head $g(\cdot)$ which is a series of three fully connected layers. 
Experiments show that a projection head improves the quality of the encoded features $h$ \cite{chen2020simple}:
\begin{align}
        z_i &= g(h_i), z_j = g(h_j)
\end{align}
The CL loss $\mathcal{L}^{i,j}$ is defined using cosine similarity on the output of the projection head: 
\begin{align}
    % \exp(sim(z_i, z_j))\tau \\
    \mathcal{L}^{i,j} &= -\log \frac{\exp\left[\simil(z_i, z_j)/\tau\right]}{\sum_{k=1}^{2N} (1 - \delta_{ik}) \exp\left[\simil(z_i, z_k)/\tau\right]}
\end{align}
% \begin{equation}
% \label{eq:loss}
%     \ell_{i,j} = -\log \frac{\exp(\mathrm{sim}(\bm z_i, \bm z_j)/\tau)}{\sum_{k=1}^{2N} \one{k \neq i}\exp(\mathrm{sim}(\bm z_i, \bm z_k)/\tau)}~,
% \end{equation}
where $\delta_{ik}$ is the Kronecker delta function which evaluates to 1 if $i=k$  and 0 otherwise. \reasat{Here, $\simil$ computes the cosine similarity score (dot product) of the representations and $\tau$ is a temperature parameter that effects the smoothness of the probability distribution of the classes. Experimentally, it has been observed that training the CL model with $\tau < 1$ accentuates differences between the scores (i.e., difference in scores among similar looking negative images will be more prominent), making the model learn from hard negative examples. \cite{chen2020simple}.}
Note that, after training is completed, we throw away the projection head $g(\cdot)$ and
use encoder $f(\cdot)$ and representation $h$ to train a proxy model.

\subsection{Proxy Model}
The proxy model $f_p(\cdot)$ is a neural network with a single fully connected layer. We extract image features using the CL model encoder and train the proxy model on these features. The output of the proxy model is used to select informative samples from the unlabeled image pool. \reasat{At the end of each contrastive learning cycle, the active learning cycle is executed. During this cycle, images are drawn from the labaled guiding samples ${S_L}$ and passed through the image encoder $f(\cdot)$ to compute image features $h$.  The proxy model $f_p(\cdot)$ is trained on these image features in binary classification setting (tumor/non-tumor). At the end of training, samples are drawn from the unlabaled data subset $U$, the feature is extracted using $f(\cdot)$, and the trained $f_p(\cdot)$ output is extracted. Output of $f(\cdot)$ or $f_p(\cdot)$ is utilized based on the chosen active sampling strategiy as described below.}

\subsection{Active Sampling}
We extract the image features  using the CL model and use active sampling strategies to select most informative samples. For this paper we use uncertainty sampling and coreset sampling.

\emph{Uncertainty Sampling.}
Uncertainty sampling is the most common sampling criteria in which an active learner selects the instances about which it is least certain how to label. The uncertain samples normally reside near the class decision boundaries. We use entropy \cite{shannon1948mathematical}, the most common measure of uncertainty: 
\begin{align}
    H(x_i)\equiv  - \sum_{j}{P(y_j\mid x_i;\theta)\log P(y_j\mid x_i;\theta)},
     \label{equ:entropy}
\end{align}
where $y_j$ ranges over all possible labels and $x_i$ is data. Entropy is an information-theoretic
measure that quantifies information content.

\emph{Coreset Sampling.}
Coreset formulates the active sampling problem as choosing a subset of samples that spans the dataset \cite{sener2017active}. The subset selection problem is formulated as a k-center problem using a greedy approximation shown in Algorithm \ref{alg:greedy}.
% using a greedy approach shown in  Algorithm~\ref{alg:greedy}. If $OPT=\min_{\mathbf{s}^1} \max_i \min_{j \in
% \mathbf{s}^1 \cup \mathbf{s}^0} \Delta(\mathbf{x}_i,\mathbf{x}_j)$, the greedy algorithm shown in
% Algorithm~\ref{alg:greedy} is proven to have a solution ($\mathbf{s}^1$) such that; $ \max_i \min_{j \in \mathbf{s}^1
% \cup \mathbf{s}^0} \Delta(\mathbf{x}_i,\mathbf{x}_j) \leq 2 \times OPT$.
% \begin{wrapfigure}{R}{0pt}
% \begin{minipage}{0.44\textwidth} 
% \vspace{-8mm}
\begin{algorithm}[h] 
   \caption{k-Center Greedy  \cite{sener2017active}} 
   \label{alg:greedy} 
   \begin{algorithmic} 
   \STATE {\bfseries Input:} data $\mathbf{x}_i$, existing unlabeled image pool $\mathbf{s}^0$ and a budget $b$ 
   \STATE Initialize $\mathbf{s}=\mathbf{s}^0$ \REPEAT \STATE $u=\arg\max_{i \in [n] \setminus \mathbf{s}} \min_{j \in \mathbf{s}} \Delta(\mathbf{x}_i, \mathbf{x}_j)$ \STATE $\mathbf{s} = \mathbf{s} \cup \{u\}$ 
   \UNTIL {$|\mathbf{s}|=b+|\mathbf{s}^0|$} 
   \STATE {\bfseries return} $\mathbf{s} \setminus \mathbf{s}^0$ \end{algorithmic}
\end{algorithm} 
% \vspace{-10mm}
% \end{minipage} 
% \end{wrapfigure}  

 \reasat{While performing coreset sampling, output of $f(\cdot)$ is utilized and for uncertainty sampling we use the output of $f_p(\cdot)$}. Uncertainty sampling focuses on samples in the decision boundary, while coreset focuses on sample diversity. These are the two most common sampling methods used in the literature. A complete list of existing strategies can be found in \cite{settles2009active}.
%There are others like Query-By-Committee, Expected Model Change, Variance Reduction and Fisher Information Ratio, Estimated Error Reduction, Density-Weighted Methods, which are not exlpored in this paper. \textbf{cite}
% Also, BALD (Houlsby et al., 2011) has been successful in
% deep learning (Gal et al., 2017; Shen et al., 2017) but is restricted to Bayesian neural networks or
% networks with dropout (Srivastava et al., 2014) as an approximation (Gal & Ghahramani, 2016).

\subsection{Proposed Framework}

\label{method_description}
Let $U^{t}$ be the unlabeled image pool at iteration $t$, $S_L$ is a small set of images with labels to train the proxy model and $S_\mathrm{test}$ is the test set. 
We sample a initial random pool of images $S^{t} \subset U^{t}$ of budget $b$ and
update the unlabeled image pool, $U^t \leftarrow U^t \setminus S^t$. The subset $S^t$ is used to train the SimCLR model.
%$g \circ f$.
% ($f(.)$ is the encoder portion of the model and $f_h$ is the projection head). 
We freeze the encoder $f(\cdot)$ and extract features $h_i$, for $x_i \in S_L$. 
A proxy model $f_p^{t}$ was trained using features $h_i$. This proxy model is used to find the next subset of images $S^{t+1}$ to be included in the CL training set, $S^{t} \leftarrow S^{t} \cup S^{t+1}$.
If uncertainty is used as sampling criterion, the entropy $H(x_i)$ is computed using outputs of $f_p^t(\cdot)$ and we select the top $b$ samples with highest uncertainty. 
If coreset is used, we select $b$ samples following the k-center greedy algorithm presented in Algorithm \ref{alg:greedy}.
This iterative process is continued for a given number of iterations $T$. 
Note that the weights of the SimCLR model are randomly initialized in the first iteration and training continues in the subsequent iterations using the learned weights in the previous iteration. The proxy network, however, is learned from random weights in each iteration.
% i.e., $S^{t+1} = \argmax_{A \subset U \setminus S^t, |A| = N_s} \sum_{a \in A} a$.
The set $S^t \cup S^{t+1}$ is used for the next iteration of CL training. This process continues for a set number of iterations $T$. The pseudocode for the algorithm of our proposed framework is presented in Algorithm \ref{alg:proposed}.

\begin{algorithm}[t]
\caption{Contrastive Learning with Active Sampling (Proposed)}
\label{alg:proposed}
\begin{algorithmic}[1]

\STATE \textbf{Input:} 
% Initial unlabeled image pool $U^{0}$,
% Initial subset for training $S^{0}$,
% Small labeled subset 
$S_L$,
% Unlabeled image pool at iteration $t$, 
$U^{t}$,
% SimCLR model 
$g\circ f$
% proxy model trained at iteration $t$, 
$f_{p}^{t}$, 
% No. of samples selected at each iteration
% $N_s$,
% No. of iterations 
$T$
\STATE \textbf{Output:} trained encoder $f(.)$
\STATE \textbf{Initialize:}\\
$S^{0} \leftarrow \text{Random sampling on} ~U^{0}$
% $U^{0} \leftarrow U^{0}\setminus S^{0}$

\STATE \FOR{$t=0$ to $T-1$}
\STATE Train $g\circ f$ using $S^t$
% \STATE $z \leftarrow f_{e}(x), x \in U^{t}$
\STATE Train $f_{p}^{t}$ using $S_L$
\STATE $U^{t} \leftarrow U^{t} \setminus S^{t}$
% \STATE $H\leftarrow \{h(f(x),f_p^{t}), x \in U^{t}\}$
\STATE $S^{t+1} \leftarrow \text{Active Sampling on}  ~U^{t}$
% \STATE \textbf{struggling here}$S^{t+1} \leftarrow \argmax_{A\subset U^t, |A| = |N_s|} \sum_{x\in A, h \in H} h$% argsort(H) in a descending order and corresponding first Ns x samples from U
\STATE $S^{t} \leftarrow S^{t} \cup S^{t+1}$

\ENDFOR
\end{algorithmic}
\end{algorithm}

\subsection{Dataset}

\begin{figure*}[htbp]
    \centering
    
    \begin{subfigure}[t]{0.2\textwidth}
        \centering
        \includegraphics[width=0.9\columnwidth]{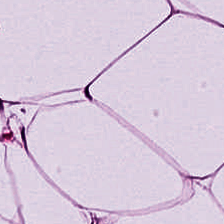}
        \caption*{Adipose (ADI).}
    \end{subfigure}%
    \begin{subfigure}[t]{0.2\textwidth}
        \centering
        \includegraphics[width=0.9\columnwidth]{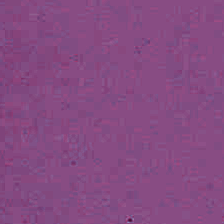}
        \caption*{Background (BACK).}
    \end{subfigure}%
    \begin{subfigure}[t]{0.2\textwidth}
        \centering
        \includegraphics[width=0.9\columnwidth]{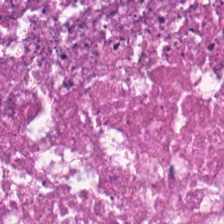}
        \caption*{Debris (DEB).}
    \end{subfigure}%
    \begin{subfigure}[t]{0.2\textwidth}
        \centering
        \includegraphics[width=0.9\columnwidth]{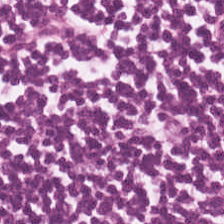}
        \caption*{Lymphocytes (LYM).}
    \end{subfigure}%
    \\
    \begin{subfigure}[t]{0.2\textwidth}
        \centering
        \includegraphics[width=0.9\columnwidth]{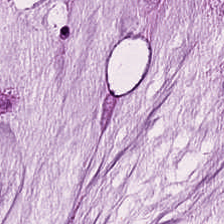}
        \caption*{Mucus (MUC).}
    \end{subfigure}%
    \begin{subfigure}[t]{0.2\textwidth}
        \centering
        \includegraphics[width=0.9\columnwidth]{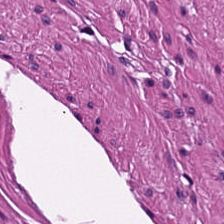}
        \caption*{Smooth Muscle (MUS).}
    \end{subfigure}%
    \begin{subfigure}[t]{0.2\textwidth}
        \centering
        \includegraphics[width=0.9\columnwidth]{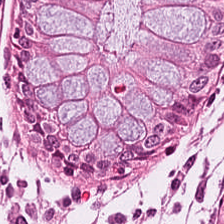}
        \caption*{Normal tissue (NORM).}
    \end{subfigure}%
    \begin{subfigure}[t]{0.2\textwidth}
        \centering
        \includegraphics[width=0.9\columnwidth]{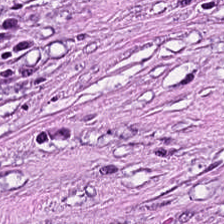}
        \caption*{Stroma (STR).}
    \end{subfigure}%
    \begin{subfigure}[t]{0.2\textwidth}
        \centering
        \includegraphics[width=0.9\columnwidth]{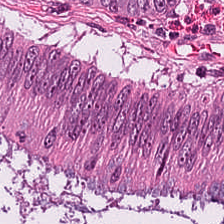}
        \caption*{Tumor (TUM).}
    \end{subfigure}%
    
\caption{The tissue types present in Kather-19 dataset.}
\label{fig:tissue_types}
\end{figure*}

% rephrase
We used the Kather-19 dataset presented in \cite{kather_jakob_nikolas_2018_1214456}. The training set consisted of $100,000$ non-overlapping image patches from hematoxylin and eosin (H\&E) stained histological images of human colorectal cancer (CRC) and normal tissue. 
Tissue classes are: adipose (ADI), background (BACK), debris (DEB), lymphocytes (LYM), mucus (MUC), smooth muscle (MUS), normal colon mucosa (NORM), cancer-associated stroma (STR), and colorectal adenocarcinoma epithelium (TUM). Samples from each class is shown in Fig.~\ref{fig:tissue_types}.
These images were manually extracted from 86 H\&E stained human cancer tissue slides from formalin-fixed, paraffin-embedded (FFPE) samples from the NCT Biobank (National Center for Tumor Diseases, Heidelberg, Germany) and the UMM pathology archive (University Medical Center Mannheim, Mannheim, Germany). Tissue samples contained CRC primary tumor slides and tumor tissue from CRC liver metastases; normal tissue classes were augmented with non-tumorous regions from gastrectomy specimen to increase variability. The test set consisted of $7180$ image patches from $50$ patients with colorectal adenocarcinoma (no overlap with patients in train set) collected from NCT tissue bank. Both the train and test set patches were $224\times 224$ pixels at 0.5 $\mu$m per pixel. All images were color normalized using Macenko's method \cite{macenko_norm}.
We created an imbalanced dataset by setting the TUM tissues as the positive class and combining all the non-tumor (NON-TUM) as negative class. The train set had $14,317$ TUM patches and $85,683$ NON-TUM patches. While the the test set had $1233$ TUM patches and $5947$ NON-TUM patches.
% \subsection{Performance}

\subsection{Experimental Details}

The CL encoder $f_{\theta}$ of the SimCLR model had a ResNet-18 architecture \cite{he2016deep} connected to a projection head that consisted of 3 fully connected layers with 128 units, \reasat{the proxy model $f_p$ was a linear layer with 512 units.} 

The model was trained using $224 \times 224$ input images. \reasat{To create two views of the same image the one of the following augmentations were randomly sampled: color jitter, random crop, random grayscale, Gaussian blur, random horizontal and vertical flip, random rotation by $\pm$ 90 deg (parameter details in \ref{app:augment}). Color jitter involves randomly altering the colors of images by adjusting brightness, contrast, hue and saturation.
Brightness augmentation adjusts the intensity of all the pixels in the image. 
% If $I$ is the original image and $\beta$ is the brightness factor, the augmented image $I'$ is given by:
% $ I' = I + \beta $
% where $\beta$ is typically a random value within a specified range, such as $[-0.5, 0.5]$.
Contrast augmentation changes the difference between the light and dark areas of the image. 
% If $\alpha$ is the contrast factor and $\mu$ is the mean intensity of the image, the augmented image $I'$ is:
% $ I' = \alpha (I - \mu) + \mu $
% where $\alpha$ is typically a random value within a specified range, such as $[0.5, 1.5]$.
Hue and Saturation augmentation modifies the intensity of the Hue and Saturation channel after converting the image to HSV (Hue, Saturation, Value) color space. The image is transformed back to RGB after the augmentation is done. Random crop randomly crops out portion of the image. Random grayscale converts the RGB image to grayscale and copies the intensities over the three channels. Gaussian blur uses a Gaussian filter to smooth (blur) the image. The flips and rotation does spatial transformation to create a different perspective of the image.
% To change the saturation, the image is converted to the HSV (Hue, Saturation, Value) color space. 
% If $S$ is the saturation channel in HSV and $\gamma$ is the saturation factor, the augmented saturation $S'$ is:
% $ S' = S \cdot \gamma $
% where $\gamma$ is typically a random value within a specified range, such as $[0.5, 1.5]$.
% Hue augmentation shifts the colors in the image along the color spectrum. In the HSV color space, if $H$ is the hue channel and $\delta$ is the hue shift factor, the augmented hue $H'$ is:
% $ H' = H + \delta $
% where $\delta$ is typically a random value within a specified range, such as $[-0.1, 0.1]$.
}
% color jitter probability 0.8, color jitter strength 0.5,    bright=color jitter strength * 0.8,
%             contrast=strength * 0.8,
%             sat=strength * 0.8,
%             hue=strength * 0.2, Minimum size of the randomized crop relative to the input size 0.08, random gray scale probability 0.2, gaussian blur float 0.5, kernel size  float 0.1, horizontal flip probability 0.5, random rotation by +90,-90 probability 0.5,  + vertical flip probability 0.5  
%             The image intensities were divided by 255. Then imagenet image mean (0.485, 0.456, 0.406) is substracted from the rgb channels and divided by the standard deviation (0.229, 0.224, 0.225).
% For MoCo, SimCLR without gaussian blur. 

The training set was randomly split into two sets: an unlabeled pool $U$ of $99,000$ images and $1000$ to train a proxy model (randomly sampled from the full train set). The original test set of size 7180 was used to evaluate the proxy model. 

As the benchmark, we trained the SimCLR model for 100 epochs using the full unlabeled pool $U$ and evaluated its feature quality by training the proxy model on the small labeled dataset. The proxy model was trained for 200 epochs, and its performance was evaluated at 20 epoch intervals on the test set.

For the proposed actively sampled framework, we set aside a subset of 1000 images (along with their labels) from $U$ and consider it as our labeled set $S_L$ from proxy model fine-tuning. The size of the subset sampled at each iteration for SSL is 1000, and the number of iterations, $T$, is 20. Note that, at each iteration, active sampling is done on a subset of 10,000 randomly selected from $U^t$ to reduce computation and data redundancy. 

Then we trained the SimCLR model iteratively following our proposed framework using subsets of images. We considered uncertainty and coreset sampling as our active sampling methods. Also, random sampling of subsets was included as an ablation study to observe the absence of active sampling. At each iteration we trained the SimCLR model for 100 epochs and the proxy model for 40 epochs. The performance of the proxy model was evaluated at the end of each iteration. For both the experiments, the SimCLR model was trained with a learning rate of 0.0001 and the proxy model with a learning rate of 0.001. The loss used for SimCLR was normalized cross-entropy with temperature ($\tau$) of 0.1,  %following suggestions from \cite{ciga2022self}
and for the classifier we used binary cross-entropy loss. For both models we used the Adam optimizer \cite{kingma2014adam} with $\beta_1=0.9$, $\beta_2=0.999$. The training batch size was 128. Each experiment was run three times, and the average performance is reported. The experiments were run using PyTorch on a Quadro RTX 6000. Details on Python libraries versions and machine specifications are reported in \ref{app:sw_hw}.

% Quadro RTX 6000 
\subsection{Performance Metrics}
The evaluation metric used was the F1 score which accounts for the imbalance in the dataset. The F1 score is the harmonic average of precision (P) and recall (R). 
\begin{align}
    F1 = \frac{2\mathrm{RP}}{\mathrm{R}+\mathrm{P}};
    R = \frac{\mathrm{TP}}{\mathrm{TP}+\mathrm{FN}},
    P = \frac{\mathrm{TP}}{\mathrm{TP}+\mathrm{FP}}
\end{align}
Here, TP is true positive, FN is false negative, FP is false positive. F1 score was chosen due to the imbalance in the positive and negative classes in the images. 
We also computed the average run time for the benchmark and proposed method to reach a certain F1 score.

\section{Results}

\label{sec_result}
\begin{figure}
\centering
\begin{subfigure}{\textwidth}
   \centering
    \includegraphics[width=0.55\textwidth]{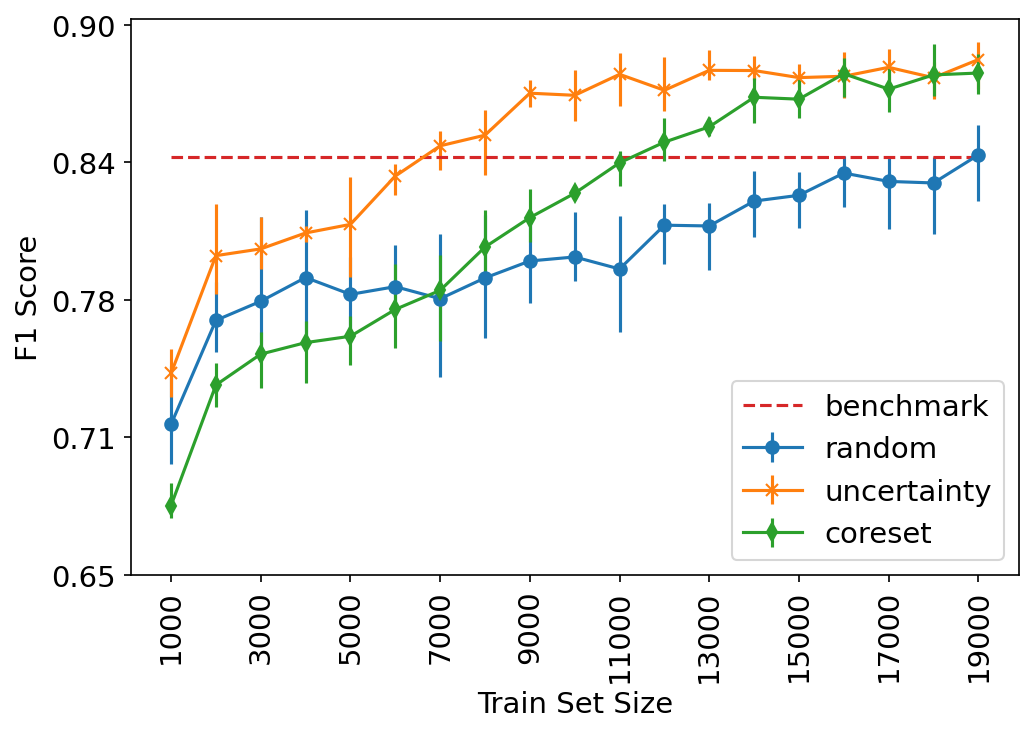}
    \caption{Progression of F1 score for different sampling strategies.} 
    \label{fig:f1_comparison}
\end{subfigure}\\
  
\begin{subfigure}{\textwidth}
   \centering
  \includegraphics[width=0.55\textwidth]{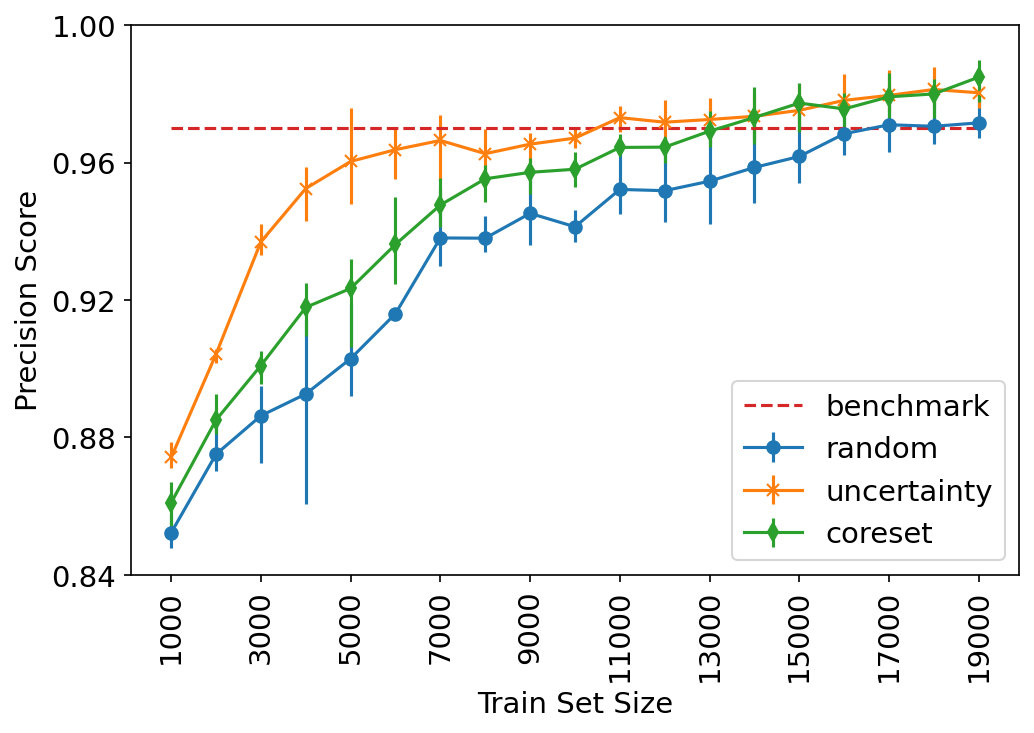}
  \caption{\reasat{Progression of precision score for different sampling strategies.}} 
  \label{fig:precision_comparison}
\end{subfigure}\\

\begin{subfigure}{\textwidth}
   \centering

  \includegraphics[width=0.55\textwidth]{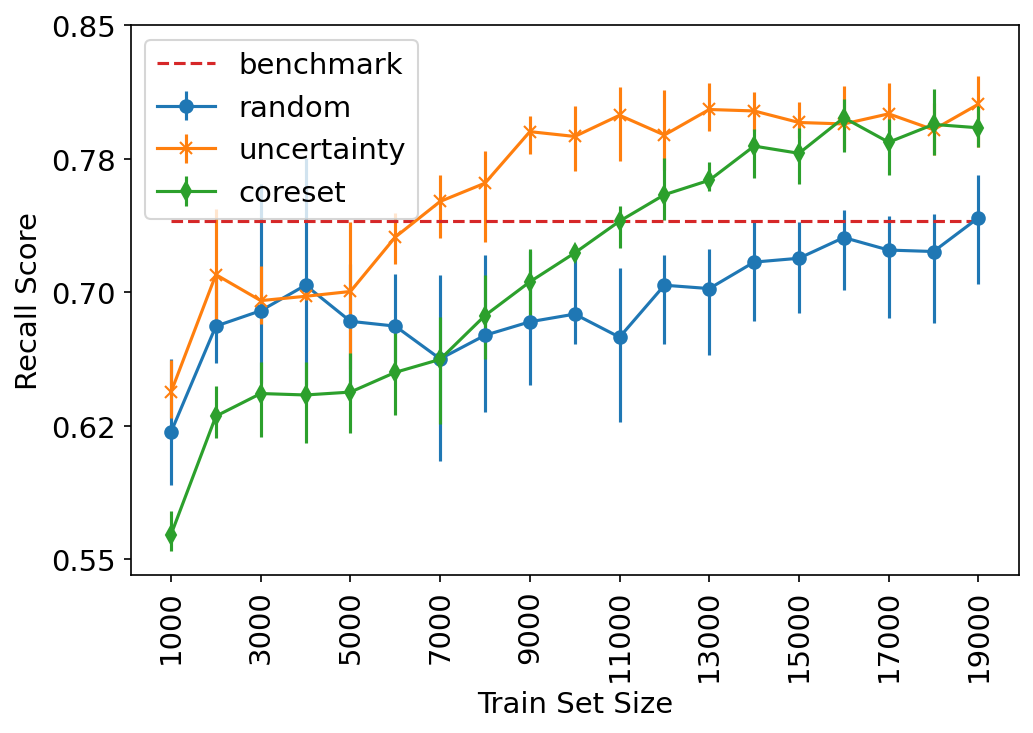}
  \caption{\reasat{Progression of recall score for different sampling strategies.}} 
  \label{fig:recall_comparison}
  \end{subfigure}

\caption{Comparison of sampling strategies. The active sampling methods (uncertainty and coreset) required less samples to reach the performance of the CL model trained on a full set of images (benchmark).} 
\label{fig:metrics_comparison}

\end{figure}

\reasat{In Fig. \ref{fig:metrics_comparison}, we observe the progression of F1, precision and recall scores through active learning iterations. The active learning based methods reach the benchmark metrics using significantly less amount of training samples}. Taking F1 score as an example, we observed a highest F1 of 0.84. To reach similar score the uncertainty and coreset based sampling strategy took 7000 and 11,000 samples respectively (Fig.~\ref{fig:f1_comparison}). Thus the sample requirement decreased by 93\% and 89\% respectively. Also, we observed that random sampling performed worse compared to the active sampling methods.
%Verified !!! 7 and 12 iterations !!!
% (622.5-228)/622 = 0.9192
% (99000-13000)/99000 = 0.8687
The benchmark achieved its highest score at the 100th epoch of SimCLR training and at the 150th epoch of the proxy model training.  The average runtime for 100 epochs of SimCLR training was 538.70 min.
% (32695.38787+31949.07714)/2/60
% (538.75-204.4)/538.75 = 0.62
To reach similar scores, the uncertainty based active sampling experiment required 7 iterations (7000 samples). The average time to train the SimCLR model for 7 iterations was 196.88 min.
Average time to train proxy model and sample selection was 7.16 min. This resulted in a total average runtime of 204.4 min.
% 11813.08926320076 + 386.16537594795227 + 43.58413100242615 = 12242.838770151138

Similarly,  k-center based sampling experiment required 11 iterations (11,000 samples). The average time to train the SimCLR model was 514.95 min.
Average time to train proxy model and sample selection was 23.70 min. This resulted in a total average runtime of 538.65 min.
% 30896.841027498245 + 687.3392248153687 + 734.5778245925903 = 32318.758076906204
Due to uncertainty sampling there was a time reduction of 62\% while coreset sampling required a similar amount of time. Further hyperparameter (such as SimCLR learning epochs, proxy model learning epochs, etc.) tuning could in the future decrease these time requirements. The runtime comparisons are summarized in Table \ref{tab:runtime}. 
% Time reduction (622.5-464)/622 = 0.15
% \usepackage{graphicx}
% \usepackage{graphicx}

% Note: It may be necessary to compile the document several times to get a multi-page table to line up properly
% \usepackage{graphicx}

\begin{table}[h]
\centering
\caption{Runtime comparison}
\label{tab:runtime}
% \resizebox{\columnwidth}{!}{%
\begin{tabular}{lrr}
\hline
Method                 & \begin{tabular}[c]{@{}c@{}}Avg Runtime\\ (min)\end{tabular} & \begin{tabular}[c]{@{}c@{}}Time Reduction\\ \%\end{tabular} \\ \hline
Benchmark              &  538.7                                                      & -                                                          \\ 
Uncertainty & 204.4                                                        & 62                                                      \\ 
Coreset     &  538.7                                                    & 0                                                       \\ \hline
\end{tabular}%
% }
\end{table}

\begin{figure}[htbp]
    \centering
    \includegraphics[width=\textwidth]{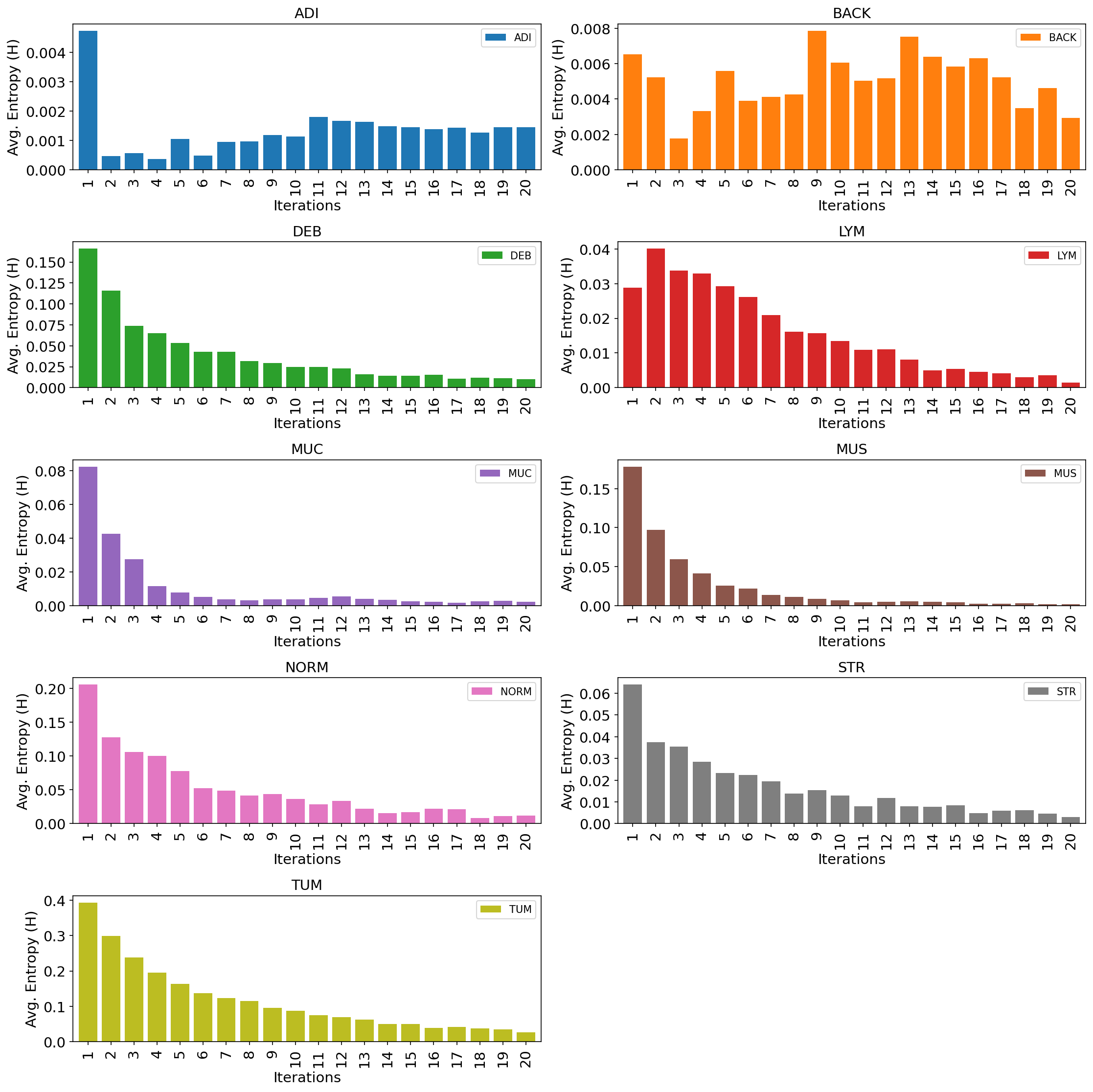}
    \caption{The change of average entropy in each iteration across different tissue types. Active sampling decreased the average entropy of the tumor samples with each iterations.}
    \label{fig:entropy_progression}
\end{figure}

\begin{figure}[htbp]
    \centering
    \includegraphics[width=\textwidth]{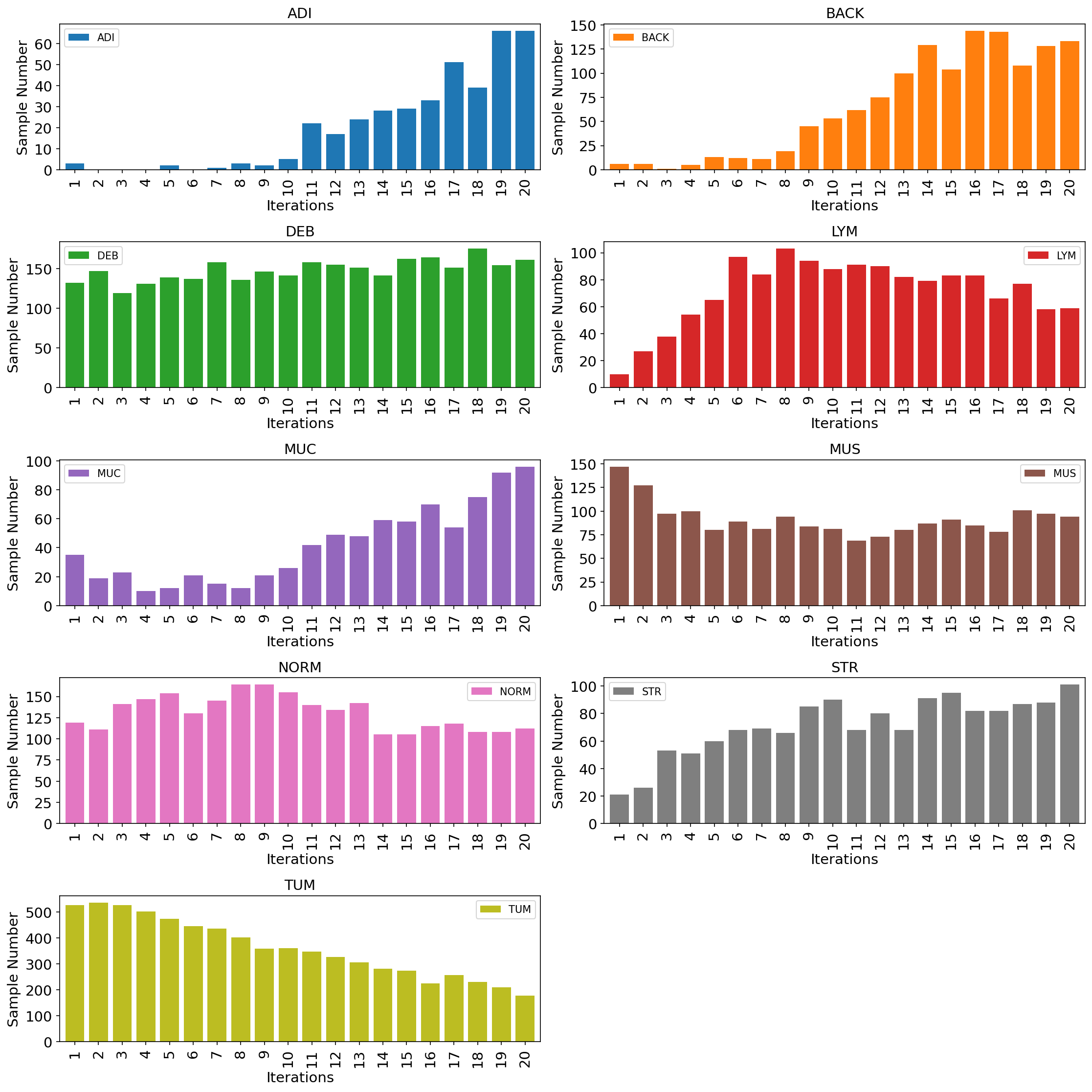}
    \caption{Number of images sampled in each iteration across different tissue types. The high average entropy led to selection of more tumor samples in the training set in the initial iterations.}
    \label{fig:class_number_progresssion}
\end{figure}

We further analyzed the uncertainty based active sampling method and looked at the change of mean entropy of the selected samples in each iteration. From Fig.~\ref{fig:entropy_progression}  we can see that, in the initial iterations, the mean entropy was highest for the tumor class. Therefore, in the first few iterations the majority of the selected samples came from the tumor class (Fig.~\ref{fig:class_number_progresssion}). Whereas, adipose and background tissue were significantly different form the tumor which induced less uncertainty to the proxy model.  
\reasat{The adipose and background classes are relatively easy to differentiate (Fig.~\ref{fig:tissue_types}), therefore their uncertainty values were extremely low to begin with (<0.01) and they were sampled less frequently in the beginning. In Fig.~\ref{fig:entropy_progression}, we saw a plateauing effect in the uncertainty for these two classes. For other classes, we saw this effect at much later iterations, as these classes have high uncertainty in the initial iterations.} In subsequent iterations, images that were more relevant to tumor classification were actively selected and added to the training pool and the uncertainty rapidly decreased and went down to the level of adipose and background classes. The CL model sampled informative tissue more frequently which resulted in a faster feature learning process.

\begin{figure*}[htbp]
    \centering
    
    \begin{subfigure}[t]{0.2\textwidth}
        \centering
        \includegraphics[width=\columnwidth]{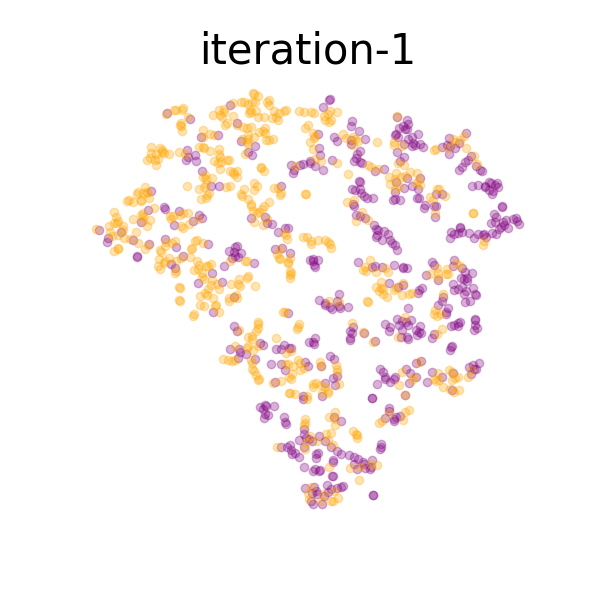}
        % \caption{All tumors.}
    \end{subfigure}%
    \begin{subfigure}[t]{0.2\textwidth}
        \centering
        \includegraphics[width=\columnwidth]{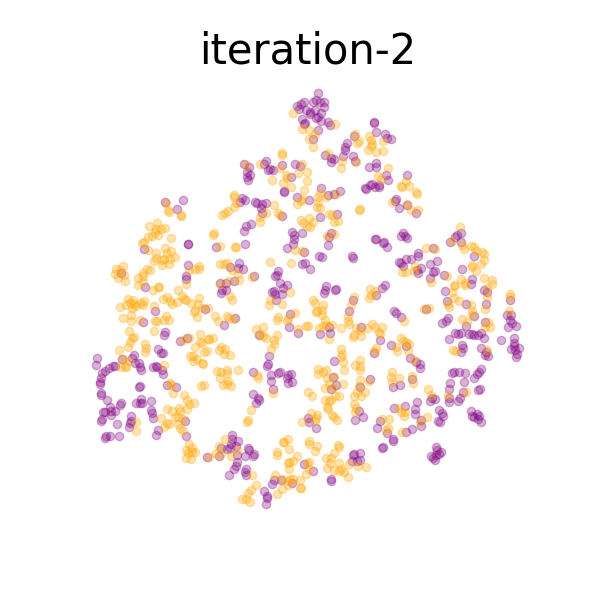}
        % \caption{All tumors.}
    \end{subfigure}%
    \begin{subfigure}[t]{0.2\textwidth}
        \centering
        \includegraphics[width=\columnwidth]{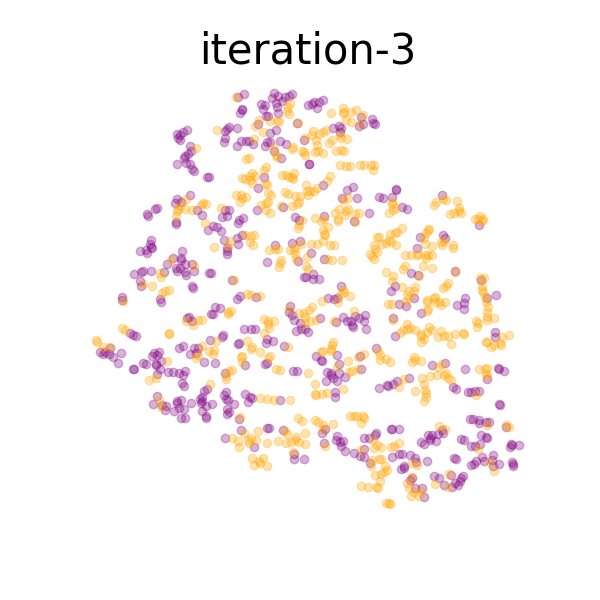}
        % \caption{All tumors.}
    \end{subfigure}%
    \begin{subfigure}[t]{0.2\textwidth}
        \centering
        \includegraphics[width=\columnwidth]{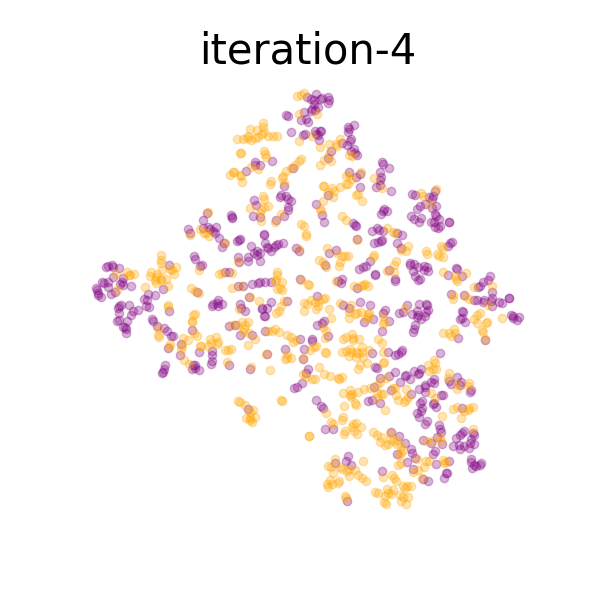}
        % \caption{All tumors.}
    \end{subfigure}%
    \begin{subfigure}[t]{0.2\textwidth}
        \centering
        \includegraphics[width=\columnwidth]{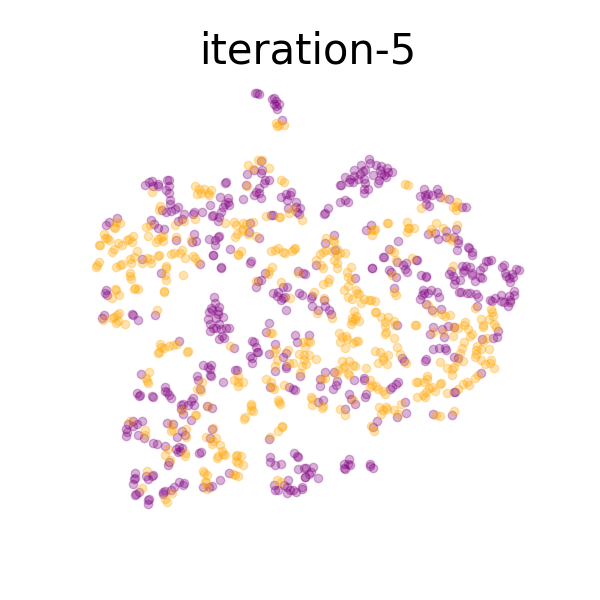}
        % \caption{All tumors.}
    \end{subfigure}
    \\
    \begin{subfigure}[t]{0.2\textwidth}
        \centering
        \includegraphics[width=\columnwidth]{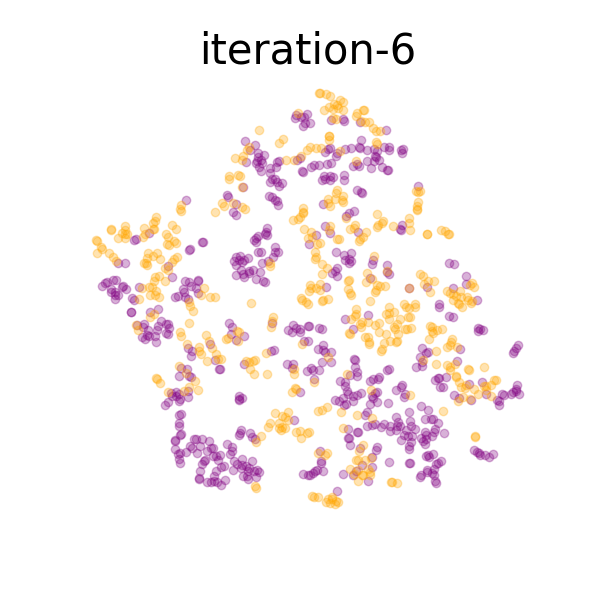}
        % \caption{All tumors.}
    \end{subfigure}%
    \begin{subfigure}[t]{0.2\textwidth}
        \centering
        \includegraphics[width=\columnwidth]{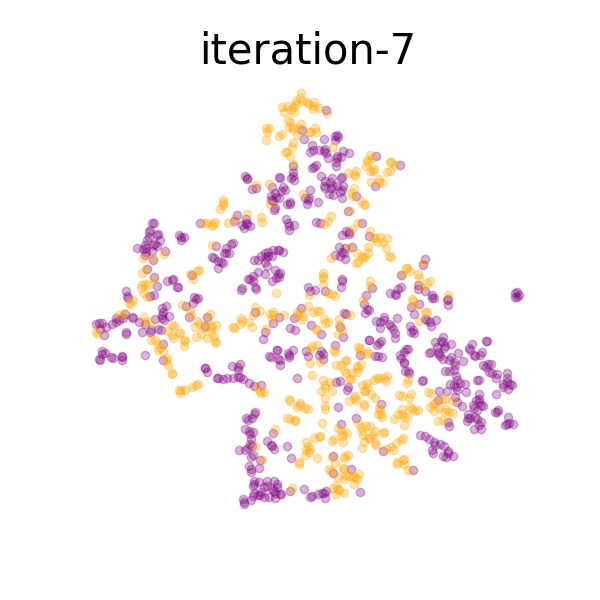}
        % \caption{All tumors.}
    \end{subfigure}%
    \begin{subfigure}[t]{0.2\textwidth}
        \centering
        \includegraphics[width=\columnwidth]{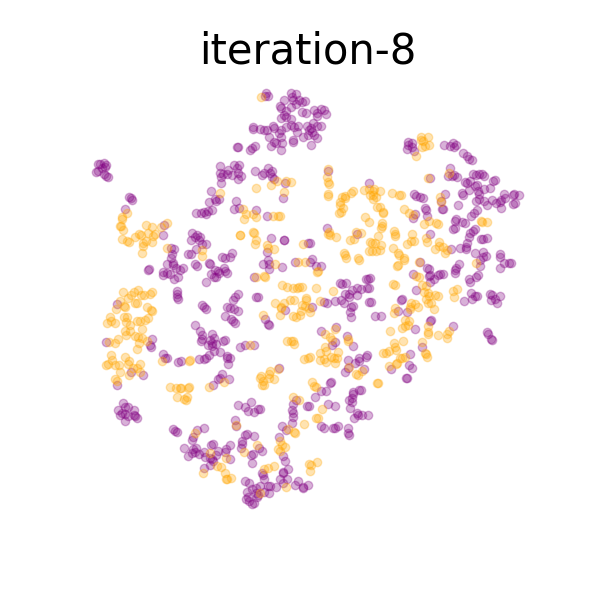}
        % \caption{All tumors.}
    \end{subfigure}%
    \begin{subfigure}[t]{0.2\textwidth}
        \centering
        \includegraphics[width=\columnwidth]{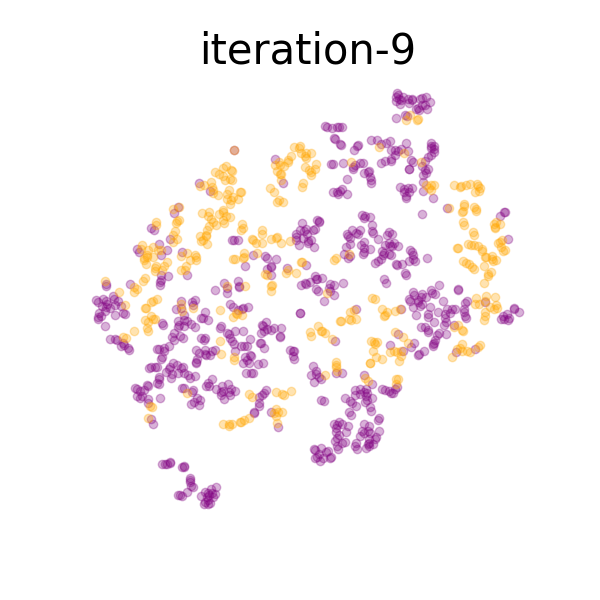}
        % \caption{All tumors.}
    \end{subfigure}%
    \begin{subfigure}[t]{0.2\textwidth}
        \centering
        \includegraphics[width=\columnwidth]{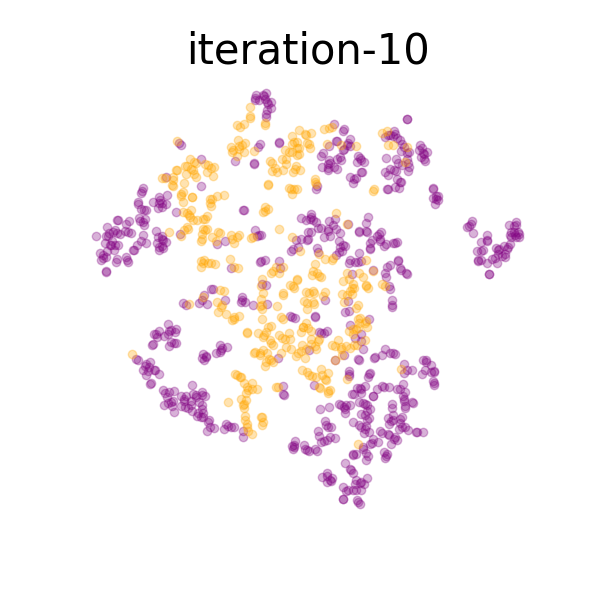}
        % \caption{All tumors.}
    \end{subfigure}%
    \\
    \begin{subfigure}[t]{0.2\textwidth}
        \centering
        \includegraphics[width=\columnwidth]{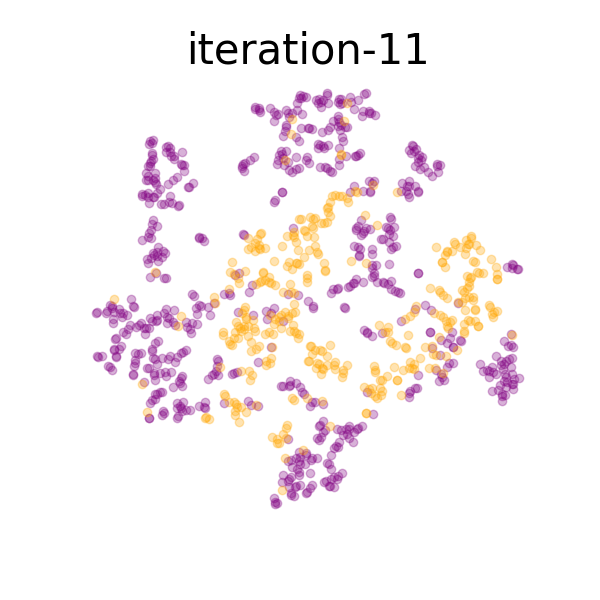}
        % \caption{All tumors.}
    \end{subfigure}%
    \begin{subfigure}[t]{0.2\textwidth}
        \centering
        \includegraphics[width=\columnwidth]{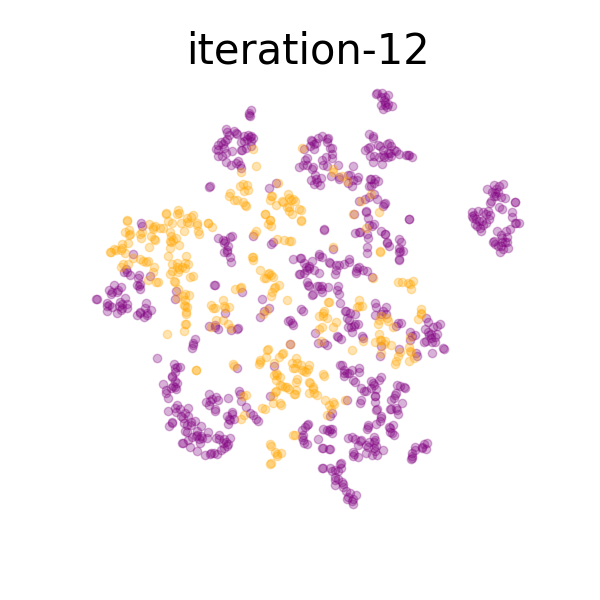}
        % \caption{All tumors.}
    \end{subfigure}%
    \begin{subfigure}[t]{0.2\textwidth}
        \centering
        \includegraphics[width=\columnwidth]{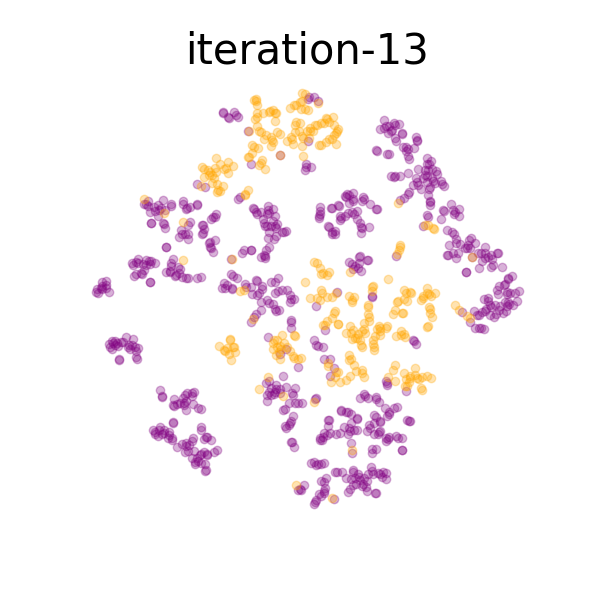}
        % \caption{All tumors.}
    \end{subfigure}%
    \begin{subfigure}[t]{0.2\textwidth}
        \centering
        \includegraphics[width=\columnwidth]{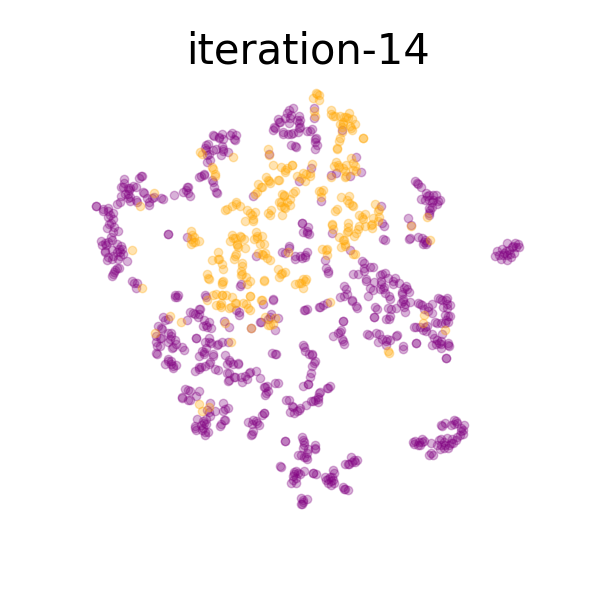}
        % \caption{All tumors.}
    \end{subfigure}%
    \begin{subfigure}[t]{0.2\textwidth}
        \centering
        \includegraphics[width=\columnwidth]{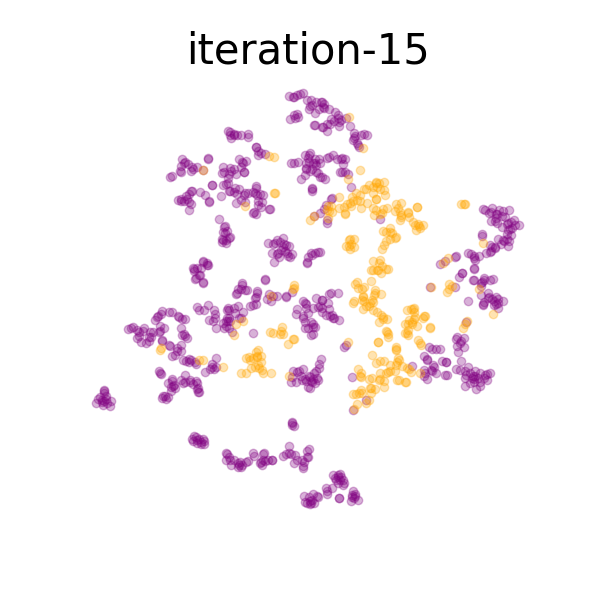}
        % \caption{All tumors.}
    \end{subfigure}%
    \\
    \begin{subfigure}[t]{0.2\textwidth}
        \centering
        \includegraphics[width=\columnwidth]{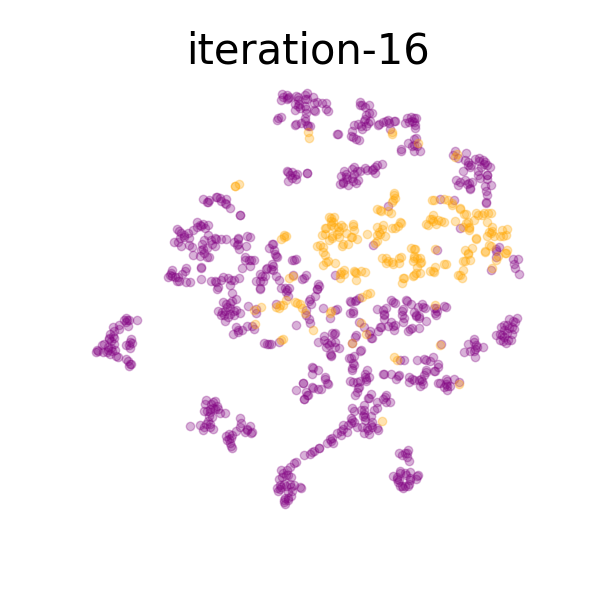}
        % \caption{All tumors.}
    \end{subfigure}%
    \begin{subfigure}[t]{0.2\textwidth}
        \centering
        \includegraphics[width=\columnwidth]{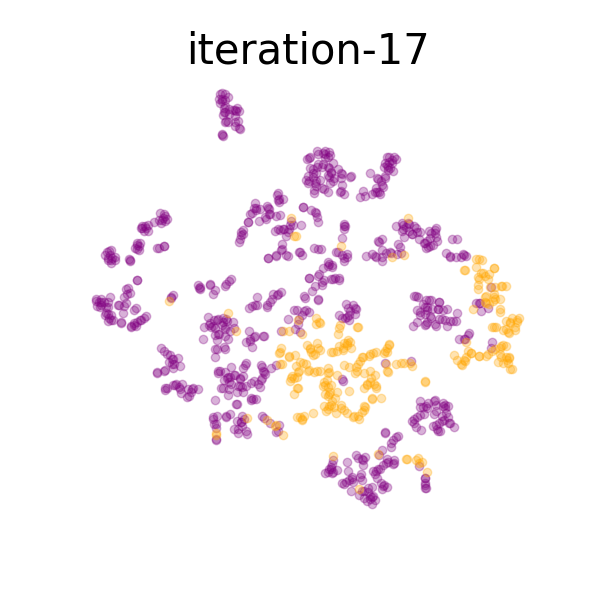}
        % \caption{All tumors.}
    \end{subfigure}%
    \begin{subfigure}[t]{0.2\textwidth}
        \centering
        \includegraphics[width=\columnwidth]{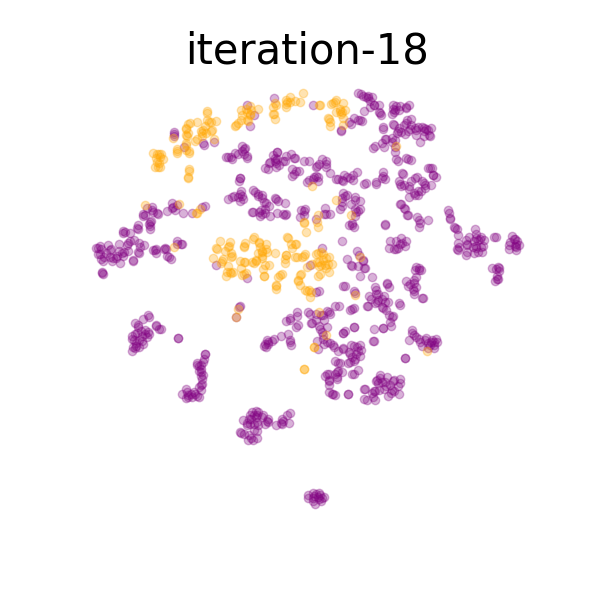}
        % \caption{All tumors.}
    \end{subfigure}%
    \begin{subfigure}[t]{0.2\textwidth}
        \centering
        \includegraphics[width=\columnwidth]{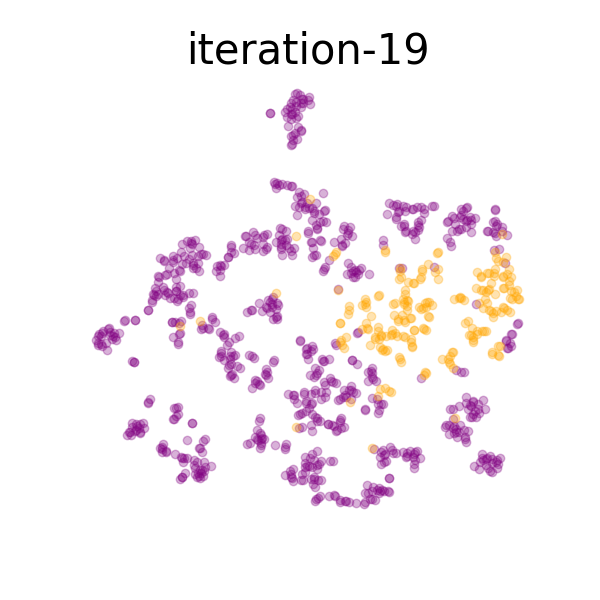}
        % \caption{All tumors.}
    \end{subfigure}%
    \begin{subfigure}[t]{0.2\textwidth}
        \centering
        \includegraphics[width=\columnwidth]{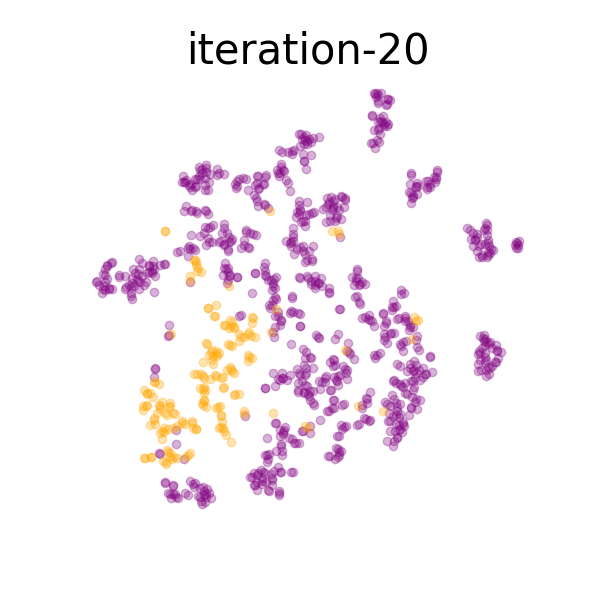}
        % \caption{All tumors.}
    \end{subfigure}%
    
\caption{t-SNE plot of the feature representations of the selected samples in each iteration. Orange (\protect\tikz \protect\draw[orange,fill=orange] (0,0) circle (.5ex);) circle represents tumor data points while purple (\protect\tikz \protect\draw[purple,fill=purple] (0,0) circle (.5ex);) data points show non-tumor samples. The feature representation of tumor samples and non-tumor samples separates gradually in each iteration.
}
\label{fig:t-sne_iteration}
\end{figure*}

In Fig.~\ref{fig:t-sne_iteration}, we show the t-SNE \cite{van2008visualizing} representation of the selected samples for different representations.  t-SNE is a stochastic neighborhood based method that compresses high-dimensional data  and facilitates the visualization of a datapoint on a two- or three-dimensional map. Here, we compressed the 512 dimensions of the image features to 2 dimensions. Only tumor and non-tumor labels are shown to better distinguish the progression of tumor tissue representation. In the first iteration, we see a lot of overlap between non-tumor and tumor representations, since, at the beginning, the model was trained on a random subset of data with low representation of tumor images. Hence, the model was unable to capture enough tumor features and separate the class in the feature space. In the next few iterations, the tumor samples increased due to active sampling, and we see the tumor class was gradually allocating a subspace in the feature space. Additionally, in the first few iterations, the tumor samples were selected more frequently compared to the last few iterations. As the tumor representation was low in the training samples in the initial stages, the uncertainty for those samples were higher, which led to increased sampling.

\begin{figure}
    \centering
    \includegraphics[width=0.65\columnwidth]{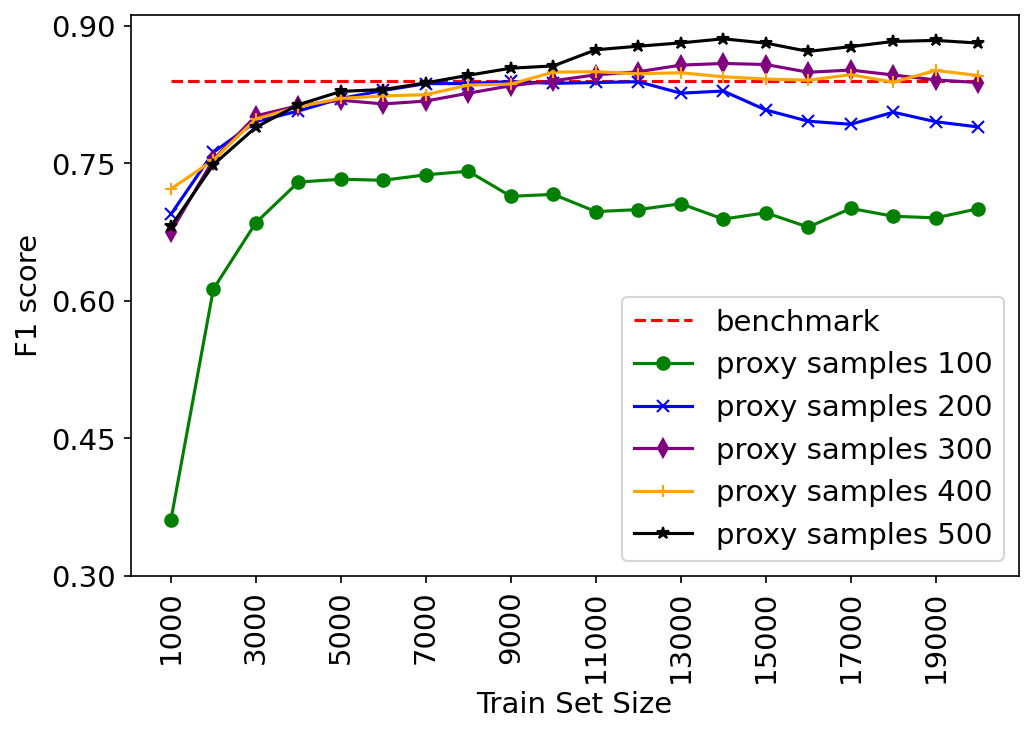}
    \caption{Effect of choosing different samples numbers ($S_L$) for the proxy models.}
    \label{fig:sample_sweep}
\end{figure}

In Fig.~\ref{fig:sample_sweep} we observe the effect of varying the number of proxy samples on the F1 score. Having a low number of samples ($<$ 300) results in lower representation of the tumor samples. This in turn created a low-quality proxy model that samples low-quality data which has a negative effect on the test score. Increasing the number of proxy samples helped, and having a sample size as low as 300 images enabled the model to reach the benchmark score.

\section{Conclusion}\label{sec_conclusion}

Active learning strategies were used in contrastive learning to speed up training by requiring less data. Also, using the uncertainty sampling, we proposed a feedback system for the contrastive learning framework to select the subset with the most amount of information related to the downstream task. For histopathological cancer classification, we achieved a speedup of 62\% while using 93\% less data. The proposed uncertainty based sampling method actively sampled tumor images. As a result the contrastively learned encoder model observed more tumor samples in batches and learned the features of tumor samples using fewer samples. This method required only a small labeled set, which has important implications for applications in which labelling is very costly, such as pathology and radiology. 

Future work could explore whether this number can be reduced and how that would affect contrastive training. While this research focused on classification as the downstream task, future research could extend this to other tasks, such as regression or segmentation. \reasat{We have only explored the application of our method in histopathology images. However, this is not a domain constrained method and we plan to explore other domains in future research.}


\appendix
% \section{Progression of Precision and Recall}

\section{Data Augmentation details}\label{app:augment}
\begin{itemize}
    \item color jitter: strength 0.5, probability 0.8
    \subitem brightness: [0, 1+strength * 0.8]
    \subitem contrast: [0, strength * 0.8]
    \subitem saturation: [0, 1+strength * 0.8]
    \subitem hue: [-strength * 0.2,+strength * 0.2]
    \item random resized crop: scale [0.08, 1], probability 0.5
    % Minimum size of the randomized crop relative to the input size 0.08
    \item random gray scale: probability 0.2
    \item gaussian blur: kernel size 0.1 * image size, probability 0.5
    \item horizontal flip: probability 0.5
    \item random rotation by +90,-90: probability 0.5,
    \item The image intensities were divided by 255. Mean of (0.485, 0.456, 0.406) was substracted from the rgb channels and divided by the standard deviation (0.229, 0.224, 0.225).
    
\end{itemize}
            
\section{Software and Hardware}
\label{app:sw_hw}
\subsection{Python Library Versions}
\label{app:lib}
\begin{itemize}
    \item \reasat{python==3.8.0}
    \item lightly==1.2.38
    \item lightly-utils==0.0.2
    \item matplotlib==3.3.0
    \item numpy==1.23.4
    \item opencv-python==4.5.5.64
    \item Pillow==9.3.0
    \item pytorch-lightning==1.5.10
    \item scikit-image==0.17.2
    \item scikit-learn==1.0.2 
    \item torch==1.9.1+cu111
    \item torchvision==0.10.1+cu111
\end{itemize}
\reasat{
\subsection{Machine Specifications}\label{app:machine}
\begin{itemize}
    \item Linux == 18.04.6 LTS (Bionic Beaver)
    \item GPU == Quadro RTX 6000 
    \item RAM == 512 GB
\end{itemize}
}

% \bibliographystyle{elsarticle-num-names} 
% \bibliographystyle{elsarticle-num} 
\bibliographystyle{elsarticle-harv} 

\bibliography{cas-refs}





% \end{thebibliography}
\end{document}